\documentclass{article}

\PassOptionsToPackage{numbers, compress}{natbib}
\usepackage[preprint]{neurips_2022}

\usepackage[utf8]{inputenc} % allow utf-8 input
\usepackage[T1]{fontenc}    % use 8-bit T1 fonts
\usepackage{hyperref}       % hyperlinks
\usepackage{url}            % simple URL typesetting
\usepackage{booktabs}       % professional-quality tables
\usepackage{amsfonts}       % blackboard math symbols
\usepackage{nicefrac}       % compact symbols for 1/2, etc.
\usepackage{microtype}      % microtypography
\usepackage{xcolor}         % colors
\newcommand{\modelname}{\emph{RGDM} }
\usepackage{caption}

\usepackage{algorithm}
\usepackage{algorithmic}
\usepackage{bm}
\usepackage{nicefrac}
\usepackage{soul}
\usepackage{braket}
\usepackage{mathtools}
\usepackage{enumitem}
\usepackage{wrapfig}
\usepackage{float}
\usepackage{multirow}

\newcommand{\ie}{\emph{i.e.} }
\newcommand{\eg}{\emph{e.g.} }

\title{Towards Controllable Diffusion Models via Reward-Guided Exploration}

\author{%
  Hengtong Zhang \\
  Tencent AI Lab\\
  \texttt{htzhang.work@gmail.com}
  \And
  Tingyang Xu \thanks{Correspondence Author.}\\
  Tencent AI Lab\\
  \texttt{tingyangxu@tencent.com} 
}

\begin{document}

\maketitle

\begin{abstract}
By formulating data samples' formation as a Markov denoising process, diffusion models achieve state-of-the-art performances in a collection of tasks. Recently, many variants of diffusion models have been proposed to enable controlled sample generation. Most of these existing methods either formulate the controlling information as an input (i.e.,: conditional representation) for the noise approximator, or introduce a pre-trained classifier in the test-phase to guide the Langevin dynamic towards the conditional goal. 
However, the former line of methods only work when the controlling information can be formulated as conditional representations, while the latter requires the pre-trained guidance classifier to be differentiable. In this paper, we propose a novel framework named \modelname (\textbf{R}eward-\textbf{G}uided \textbf{D}iffusion \textbf{M}odel) that guides the training-phase of diffusion models via reinforcement learning (RL). 
The proposed training framework bridges the objective of weighted log-likelihood and maximum entropy RL, which enables calculating policy gradients via samples from a pay-off distribution proportional to exponential scaled rewards, rather than from policies themselves.
Such a framework alleviates the high gradient variances and enables diffusion models to explore for highly rewarded samples in the reverse process.
Experiments on 3D shape and molecule generation tasks show significant improvements over existing conditional diffusion models. 
\end{abstract}

\section{Introduction}

Diffusion models have already shown their great success in density estimation~\cite{ho2020denoising,song2019estimatinggradients,nichol2021improved,kingma2021variational}, image synthesis~\cite{rombach2022high,ho2022cascaded}, 3D shape generation~\cite{luo2021diffusion,zhou20213d}, audio synthesis~\cite{chen2020wavegrad,kong2020diffwave} and super-resolution~\cite{saharia2022image}. 
This series of models define a Markov diffusion process that gradually adds random noise to data samples and then learns a reversed process to denoise the perturbations added in the diffusion process to reconstruct data samples from the noise. 
Ho et al.~\cite{ho2020denoising} showed that diffusion models essentially learn gradients of data distribution density, which is equivalent to score-based generative models like ~\cite{song2019estimatinggradients}.
Recently, a collection of literature~\cite{luo2021diffusion,ho2020denoising,rombach2022high,dhariwal2021diffusion,ho2022classifier,saharia2022palette,choi2021ilvr,song2021solving} proposes multiple variants of diffusion models to enable more precise control of generation results. Such controlled models directly benefit a collection of commercial applications, such as STABLE Diffusion\footnote{https://huggingface.co/spaces/stabilityai/stable-diffusion} and ERNIE-ViLG\footnote{https://huggingface.co/spaces/PaddlePaddle/ERNIE-ViLG}.

Based on the methodology, existing conditional diffusion models can be categorized into three types. The first type of works~\cite{luo2021diffusion,ho2020denoising,ho2022classifier,rombach2022high} directly introduces conditional variables to construct conditional noise estimator for diffusion models. However, this line of methods can only be applied when conditional information can be formulated as representation variables. 
The second type of works~\cite{dhariwal2021diffusion} manipulates the generation results by introducing pre-trained classifiers~\cite{dhariwal2021diffusion}. Nevertheless,~\cite{dhariwal2021diffusion} requires the pre-trained classifier to be differentiable. Conditional guidance from regression models or non-differentiable classifiers, such as random forest, cannot be used under~\cite{dhariwal2021diffusion}.
Thus, its application scope is also limited.
Finally, there are also some works design task-specific~\cite{saharia2022palette,choi2021ilvr,song2021solving} (\eg image-to-image translation or linear inverse imaging) conditional generation models. These methods typically require additional reference such as images, and can hardly precisely decide the generation outcome.

To tackle the drawbacks, we propose to guide the reversed process of a diffusion model via a reinforcement learning (RL) reward function for flexible and controllable generation. This is because the diffusion/reversed process in diffusion models and the Markov decision process (MDP) in RL both follow the Markov property.
However, directly applying classic RL algorithms (\eg policy gradient~\cite{sutton1999policy}, Q-learning~\cite{mnih2013playing}) to train reward-guided diffusion models can lead to performance and efficiency issues. The reasons are as follows. First, unlike autonomous controlling, the length of reversed processes can be as long as hundreds of steps. Thus, the estimated gradient given such long episodes faces high variance. Moreover, in classic RL algorithms, gradients are calculated via episodes sampled directly from the policies themselves, which is costly and non-station.
Last but not least, for some non-smooth reward functions, it is difficult for diffusion models to find any highly rewarded samples. 
 Due to these discrepancies, there is still a noticeable gap before applying RL techniques to diffusion models in real practice.

In this paper we develop a novel reinforced training framework named \modelname for diffusion models. Specifically, the proposed \modelname draws intermediate samples in the reversed process from a reward-aware pay-off distribution instead of the estimated diffusion model; and  utilizes these samples to compute policy gradients and update the diffusion model. Compared with traditional RL methods~\cite{sutton1999policy,mnih2013playing,mnih2016asynchronous}, which rely on samples from the policy (model) to calculate gradients, the proposed framework not only reduces the variance of estimated gradients but also avoids expensive sampling from the non-stationary policy (model). The theoretical analysis in this paper reveals that sampling from a stationary pay-off distribution enjoys an identical optimal point to maximum entropy RL~\cite{peters2007reinforcement,norouzi2016reward,kappen2012optimal}.

Our contribution is three-fold. First, we recognize the limitations of existing techniques that aim to guide diffusion models. Second, we propose a practical reward-guided training framework, which enjoys nice theoretical properties, to enable flexible and efficient conditional manipulation for diffusion models. Finally, through extensive experiments on two generation tasks, we demonstrate the proposed framework outperforms baselines by a clear margin.

%% math
\newcommand{\cY}{\mathcal{Y}}
\newcommand{\cX}{\mathcal{X}}
\newcommand{\en}{\mathcal{E}}
\newcommand{\be}{\mathbf{e}}
\newcommand{\by}{\mathbf{y}}
\newcommand{\bx}{\mathbf{x}}
\newcommand{\bX}{\mathbf{X}}
\newcommand{\bM}{\mathbf{M}}

\newcommand{\bz}{\mathbf{z}}
\newcommand{\bu}{\mathbf{u}}

\newcommand{\bh}{\mathbf{h}}
\newcommand{\bo}{\mathbf{o}}
\newcommand{\bs}{\mathbf{s}}
\newcommand{\ba}{\mathbf{a}}
\newcommand{\bmu}{\bm{\mu}}
\newcommand{\beps}{\bm{\epsilon}}

\newcommand{\ot}{\bo_t}
\newcommand{\at}{\ba_t}
\newcommand{\op}{\mathcal{O}}
\newcommand{\opt}{\op_t}
\newcommand{\kl}{D_\text{KL}}
\newcommand{\tv}{D_\text{TV}}
\newcommand{\ent}{\mathcal{H}}

\newcommand{\bzhi}{\bz^\text{hi}}

\newcommand{\E}{\mathbb{E}}

\newcommand{\model}{\emph{ROR }}

%% editing comment
\newcommand{\cmt}[1]{{\footnotesize\textcolor{red}{#1}}}
\newcommand{\cmto}[1]{{\footnotesize\textcolor{orange}{#1}}}
\newcommand{\note}[1]{\cmt{Note: #1}}
\newcommand{\question}[1]{\cmto{Question: #1}}
\newcommand{\sergey}[1]{{\footnotesize\textcolor{blue}{Sergey: #1}}}
\newcommand{\ak}[1]{{\footnotesize\textcolor{red}{AK: #1}}}
\newcommand{\edits}[1]{\textcolor{red}{#1}}

%% abbreviations
\newcommand{\x}{\mathbf{x}}
\newcommand{\z}{\mathbf{z}}
\newcommand{\y}{\mathbf{y}}
\newcommand{\w}{\mathbf{w}}
\newcommand{\data}{D}

\newcommand{\nth}{\text{th}}
\newcommand{\pr}{^\prime}
\newcommand{\tr}{^\mathrm{T}}
\newcommand{\inv}{^{-1}}
\newcommand{\pinv}{^{\dagger}}
\newcommand{\real}{\mathbb{R}}
\newcommand{\gauss}{\mathcal{N}}
\newcommand{\norm}[1]{\left|#1\right|}
\newcommand{\trace}{\text{tr}}

%% specifics for the paper
\newcommand{\reward}{r}
\newcommand{\policy}{\pi}
\newcommand{\mdp}{\mathcal{M}}
\newcommand{\states}{\mathcal{S}}
\newcommand{\actions}{\mathcal{A}}
\newcommand{\observations}{\mathcal{O}}
\newcommand{\transitions}{T}
\newcommand{\initstate}{d_0}
\newcommand{\freq}{d}
\newcommand{\obsfunc}{E}
\newcommand{\initial}{\mathcal{I}}
\newcommand{\horizon}{H}
\newcommand{\rewardevent}{R}
\newcommand{\probr}{p_\rewardevent}
\newcommand{\metareward}{\bar{\reward}}
\newcommand{\discount}{\gamma}
\newcommand{\behavior}{{\pi_\beta}}
\newcommand{\bellman}{\mathcal{B}}
\newcommand{\qparams}{\phi}
\newcommand{\qparamset}{\Phi}
\newcommand{\qset}{\mathcal{Q}}
\newcommand{\batch}{B}
\newcommand{\qfeat}{\mathbf{f}}
\newcommand{\Qfeat}{\mathbf{F}}
\newcommand{\feasible}{\mathcal{F}}
\newcommand{\I}{\mathbb{I}}

\newcommand{\poison}{\mathcal{P}}
\newcommand{\distribution}{\mathcal{D}}

\newcommand{\traj}{\tau}

\newcommand{\pihi}{\pi^{\text{hi}}}
\newcommand{\pilo}{\pi^{\text{lo}}}
\newcommand{\ah}{\mathbf{w}}

\newcommand{\proj}{\Pi}

\newcommand{\loss}{\mathcal{L}}
\newcommand{\eye}{\mathbf{I}}

\newcommand{\pimix}{\pi_{\text{mix}}}

\newcommand{\pib}{\bar{\pi}}
\newcommand{\epspi}{\epsilon_{\pi}}
\newcommand{\epsmodel}{\epsilon_{m}}
\newcommand{\epss}{\epsilon_{s}}
\newcommand{\epsa}{\epsilon_{a}}
\newcommand{\vdelta}{\bm{\delta}}

\newcommand{\return}{\mathcal{R}}

\newcommand{\conpen}{\mathcal{C}}

%% math
% \newcommand{\cY}{\mathcal{Y}}
% \newcommand{\cX}{\mathcal{X}}
% \newcommand{\en}{\mathcal{E}}
% \newcommand{\be}{\mathbf{e}}
% \newcommand{\by}{\mathbf{y}}
% \newcommand{\bx}{\mathbf{x}}
% \newcommand{\bz}{\mathbf{z}}
% \newcommand{\bo}{\mathbf{o}}
% \newcommand{\bs}{\mathbf{s}}
% \newcommand{\ba}{\mathbf{a}}
% \newcommand{\ot}{\bo_t}
% \newcommand{\st}{\bs_t}
% \newcommand{\at}{\ba_t}
% \newcommand{\op}{\mathcal{O}}
% \newcommand{\opt}{\op_t}
% \newcommand{\kl}{D_\text{KL}}
% \newcommand{\tv}{D_\text{TV}}
% \newcommand{\ent}{\mathcal{H}}

% \newcommand{\bzhi}{\bz^\text{hi}}

% \newcommand{\E}{\mathbb{E}}
\section{Related Works}\label{sec:related}

\paragraph{Conditional Diffusion Models}
A collection of diffusion models has been designed for conditional generation. Based on the methodology, existing work can be classified into three categories. The first line of methods~\cite{ho2020denoising,luo2021diffusion,rombach2022high} introduce conditional variables and construct conditional noise approximators (or score approximators) for diffusion models. Early works either directly introduce condition variables as an input of the noise approximators~\cite{ho2020denoising}, or introduce an encoder~\cite{luo2021diffusion} to encode more complex conditional knowledge (e.g., reference image or text). Later, works like \cite{rombach2022high} first project samples from image space to low-dimensional latent space and then perform diffusion models on the latent space. In each reversed step, the conditional representation is mapped to the intermediate layers of the noise approximators via a cross-attention layer.
\cite{ho2022classifier} jointly train a conditional and an unconditional diffusion model, and we combine the resulting conditional, and unconditional score estimates to attain a trade-off between sample quality and diversity similar to that obtained using classifier guidance.
The second line of methods~\cite{dhariwal2021diffusion} manipulates the sampling phase of diffusion models to guide a trained model to generate samples that satisfy certain requirements. For instance, \cite{dhariwal2021diffusion} proposes to utilize a pre-trained classifier $p(y \mid x_t)$ to guide the denoising model towards generating images with the attribute $y$. 

Finally, there are also some works~\cite{saharia2022palette,choi2021ilvr,song2021solving} design task-specific conditional generation methods for image-to-image translation and linear inverse imaging problems. For example, Saharia et al.~\cite{saharia2022palette} directly injects a corrupted reference image into the noise approximators of diffusion models for image-to-image translation.
Choi et al. \cite{choi2021ilvr} proposes a method upon unconditional DDPM. Particularly, it introduces a reference image to influence the vanilla generation process. In each transition of the reversed process, the intermediate denoising result is synthesized with the  corrupted reference image to let the reference image influence the generation result. 
{Unlike} the existing conditional diffusion model, this paper proposes a flexible reinforced framework that utilizes a pre-trained classifier/regressor to guide the diffusion model towards the desired condition \underline{in the training phase}. To our knowledge, this is the first training-phase conditional diffusion framework alternative to directly introducing condition variables.

\paragraph{3D Point Cloud Generation}
This paper is also related to 3D shape generation since the experiments are carried out on three point cloud datasets. Shape generation via deep model is a fundamental topic in computer vision.
Unlike images, 3D shapes can be represented as: voxel grids, point clouds and meshes, etc.
In this paper, we focus on generating 3D shapes in the form of point clouds from scratch. Most existing works in this domain can be roughly classified into:
\textit{autoregressive-based}~\cite{sun2020pointgrow} (learn the joint distribution of 3D point coordinates, and sample points one-by-one in the generation phase), 
\textit{flow-based}~\cite{yang2019pointflow,mittal2022autosdf,klokov2020discrete,cai2020learning} (learn a sequence of invertible transformations of points that can transform real data to a simple distribution or vice-versa.)
and \textit{GAN-based}~\cite{shu20193d,li2021sp,ramasinghe2020spectral,tang2022warpinggan} (simultaneously learn the generator and the discriminator via a mini-max game). 
Recently, Luo et al.~\cite{luo2021diffusion} adopts a diffusion model to learn a Markov reverse diffusion process for point clouds given a latent distribution. Similar idea is also used in existing score-based approaches, such as ShapeGF~\cite{cai2020learning}.  Essentially, these two methods learn the gradient density of data distribution and gradually move points along gradients. \textit{We have included~\cite{luo2021diffusion} in our experiment as a comparison method.} 
\section{Background: Diffusion Models}\label{sec:background}

Before heading to the proposed \emph{reward-guided diffusion model}, let us briefly review the general formulation of diffusion models. 

Considering a data sample  $\bX_0$, diffusion models such as~\cite{ho2020denoising,nichol2021improved,kingma2021variational,sohldickstein2015diffusion} are inspired by non-equilibrium thermodynamics, in which data points gradually diffuse into chaos. A diffusion model consists of a forward {\em diffusion process} and a learnable backward {\em reversed process}.

\paragraph{Diffusion Process} In the diffusion process, multivariate Gaussian noise is added to the sample step-by-step which is Markovian. The transition distribution from step $t-1$ to step $t$ is formulated as:
\begin{equation}
	q(\bX_{t} \mid \bX_{t-1}) = \gauss(\bX_{t} ; \alpha_{t} \bX_{t-1}, \beta_{t} \eye),
	\label{eq:markov}
\end{equation}
where $\bX_{t}$ denotes the noisy intermediate sample in step $t$ and $\eye$ is an identity matrix. 
Two sets of noise schedule parameters, $\alpha_{t}$ and $\beta_{t}$, control how much signal is retained, and how much noise is added, respectively. 
Those parameters are either formulated in fixed form~\cite{sohldickstein2015diffusion,ho2020denoising} or learned via neural networks~\cite{kingma2021variational}. In this paper, we adopt \textit{variance-preserving} strategy as used by~\cite{ho2020denoising,sohldickstein2015diffusion} to let $\alpha_t=1-\beta_t$.

\paragraph{Reversed Process} The reversed process, which is viewed as the reverse of the diffusion process, aims to generate meaningful samples from random noise. In such a process, the random noise reversely passes through the Markov chain in Eq.~\eqref{eq:markov} and recovers the desired sample.

Suppose we know the exact reverse distribution as $q(\bX_{t-1} \mid \bX_{t}, \bX_0 )$, 
we can reversely run the diffusion process to get a denoised sample~\cite{ho2020denoising,kingma2021variational,nichol2021improved}. 
However, $q(\bX_{t-1} \mid \bX_{t}, \bX_0 )$ is inducted based on the entire data distribution. Hence, we approximate it
using a neural network:
\begin{align}
	p_\theta(\bX_{t-1} \mid \bX_{t}) & = (\bX_{t-1} \mid \bmu_\theta( \bX_{t}, t), \eta_t \eye), 
	\label{eq:p_theta}
\end{align}
where $\eta_t$ is the scheduled variance at step $t$ and we set $\eta_t = \beta_t$. $\bmu_\theta( \bX_{t}, t)$ is a learnable estimator for the mean of $\bX_{t-1}$ w.r.t $\bX_t$ and $t$. 
% i.e. the same as q(zs \mid zt, x), but with the original data x replaced by the output of a denoising model xˆθ(zt;t) that predicts x from its noisy version zt
For better description, we postpone the detailed implementation of $\bmu_\theta( \bX_{t}, t )$ and $\eta_t$ in \emph{the next paragraph}.

\paragraph{Training}
The training of diffusion models is performed by minimizing the variational lower bound on negative log likelihood:
\begin{equation}
\begin{split}
    \loss_{MLE} = & \; \E ( - \log p(\bX_0) ) \\
    \leq & \; \underbrace{ \kl (q(\bX_{T} \mid \bX_0) \mid\mid  p(\bX_{T})) }_{\loss_T}  +  \sum_{t=1}^T \underbrace{ \kl ( q(\bX_{t-1} \mid \bX_{t}, \bX_0) \mid\mid  p_\theta(\bX_{t-1} \mid \bX_{t}) ) }_{ \loss_{ll,t} }. 
\end{split}
\label{eq:loss}
\end{equation}
Here, $\loss_{ll,t}$ suggests that the estimated reverse distribution $ p_\theta(\bX_{t-1} \mid \bX_{t}) $ should be close to the exact reverse distribution $q(\bX_{t-1} \mid \bX_{t}, \bX_0 )$. $\loss_{T}$ characterizes the deviation between $ q(\bX_{T} \mid \bX_0) $ and a standard Gaussian $ p(\bX_{T}) $. 
Here $q(\bX_{t-1} \mid \bX_{t}, \bX_0)$ is formulated as:
\begin{equation}
	\begin{split}
		q(\bX_{t-1} & \mid \bX_{t}, \bX_0) = \gauss(\bX_{t-1} \mid \bmu_t(\bX_{t}, \bX_0), \gamma_t \eye), 
	\end{split}
	\label{eq:qx_t-1|x,t_x*}
\end{equation}
where $ \gamma_t = \frac{\beta_t (1 - \alpha_{t-1}) }{ \bar{\alpha}_t }$. and 
%\begin{equation}
$\bmu_t(\bX_{t}, \bX_0)  =  \frac{ \sqrt{\alpha_t} ( 1 - \bar{\alpha}_{t-1} ) }{ 1 - \bar{\alpha}_{t} } \bX_{t} + \frac{ \beta_t \sqrt{ \bar{\alpha}_{t-1} }  }{ 1 - \bar{\alpha}_t } \bX_0$ 
%\end{equation}
with $ \bar{\alpha}_t = \prod_{m=1}^{t} \alpha_m $.
Regarding \cite{ho2020denoising}, there are ways to parameterize $\mu_\theta(\bX_{t}, t)$ in
Eq.~\eqref{eq:p_theta} as 
%\begin{equation}
$\bmu_\theta(\bX_{t}, t) =  \frac{1}{ \sqrt{ \alpha_t} } \bX_{t} - \frac{\beta_t  }{ \sqrt{ \alpha_t (1 - \bar{\alpha}_t ) } } \beps_\theta (\bX_{t}, t)$.
With both Eq.~\eqref{eq:p_theta} and Eq.~\eqref{eq:qx_t-1|x,t_x*} in Gaussian, one can write the KL divergence of $\loss_{ll, t}$ in closed-form, as suggested in \cite{ho2020denoising} and \cite{kingma2021variational}:
\begin{equation}
	\loss_{ll, t} = \E_{q} \left[ \frac{1}{2\eta_t} || \bmu_t( \bX_{t}, \bX_0) - \bmu_\theta( \bX_{t}, t )|| ^2 \right ]= \E_{q}  \frac{1}{2\eta_t}  ||  \beps - \beps_\theta (\bX_{t}, t) || ^2.
	\label{eq:loss_ll}
\end{equation}

\paragraph{Sampling} The sampling process resembles Langevin dynamics with the estimator of data density gradients, 
i.e., $\beps_\theta$. Concretely, the sampling process generates intermediate samples iteratively via:
\begin{equation}
\widehat{\bX}_{t-1} = \frac{1}{ \sqrt{ \alpha_t} } \bX_{t} - \frac{\beta_t \beps_\theta (\bX_{t}, t) }{ \sqrt{ \alpha_t (1 - \bar{\alpha}_t ) } }  + \sqrt{ \beta_t } \beps,  \forall t=T,\cdots, 1.
	\label{eq:sampling}
\end{equation}

\section{\modelname: Reward Guided Diffusion Model} \label{sec:method}

\begin{figure*}
    \centering
    \includegraphics[width=\textwidth]{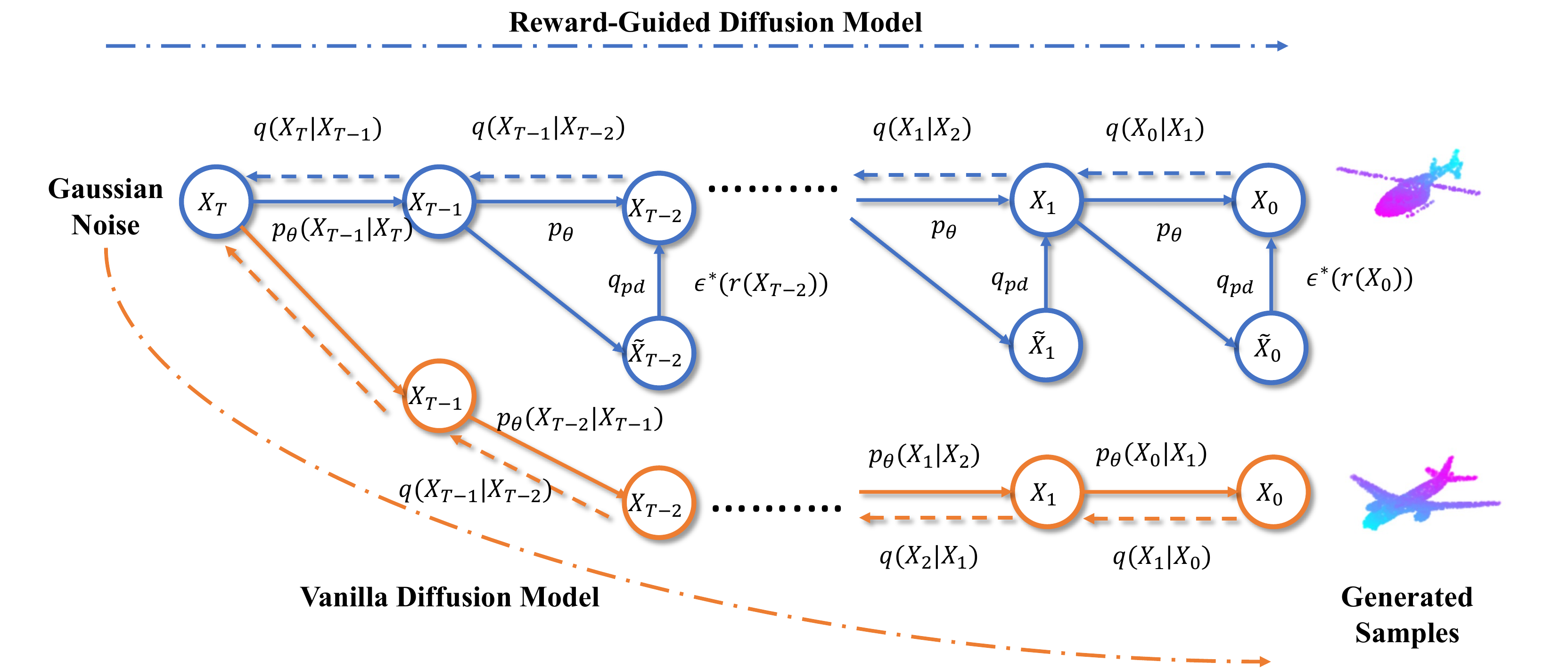}
    \caption{The directed graphical model of the proposed reward-guided diffusion model \modelname. The orange and  blue trajectory denote the diffusion/reverse process of vanilla DDPM and  \modelname, respectively.}
    \label{fig:model}
\end{figure*}

\subsection{Overview}

% A vast majority of controlled generative models in principle learn conditional distributions of the form $p(\bz \mid \by)$,  where $\bz$ and $\by$ stands for  latent representations and the conditional inputs, respectively.  Particularly for diffusion model, this can be implemented as a conditional noise estimator $\beps_\theta(\bX_t, t,\by)$~\cite{luo2021diffusion,ho2020denoising} (unlike unconditional estimator $\beps_\theta(\bX_t, t)$).   Nevertheless, these controlled generative models can only be applied when conditional information can be formulated as representation variables. 
% To circumvent this major drawback, we  propose an alternative controllable training framework for diffusion models based on reinforcement learning (RL).

Applying RL to diffusion model training is quite intuitive, since diffusion models 
perform generation via iterative refinements, which can be viewed as a Markov multi-step decision process in the context of RL. 
Concretely, given a dataset $\data$, we compute $\E_\data ( \reward(\bX) )$ as a measure of the empirical reward, which evaluates the fulfillment of the controllable generation goal, and hopes to maximize empirical rewards during training.  

In this paper, we maximize the expected reward with a maximum entropy regularizer~\cite{haarnoja2018acquiring}, given as: $\max_\pi \E_{ \data} \sum_t [r_t + \ent(\pi (\cdot | s_t))] $, where $r$, $s$, $\pi$, $\ent$ denote reward, state, policy and entropy, respectively. 
By treating $ p_\theta( \bX_{t-1} \mid \bX_{t} )$ in diffusion models as RL policy, one can adopt such a learning objective to train reward-guided diffusion models:
\begin{equation}
\begin{split}
\loss_{RL} = & - \E_{ \data} 
\bigg\{ 
\sum_{t=1}^T  \big[  \ent( p_\theta( \bX_{t-1} \mid \bX_{t} ) ) + \E_{ p_\theta(\bX_{t-1} \mid \bX_{t} ) } \reward( \bX_{t-1} )  \big]  \\
&+ \left[  \ent( p_\theta( \bX_{0} ) ) + \E_{ p_\theta(\bX_{0} ) }   \reward( \bX_{0} )  \right]
\bigg\},
\end{split}
\label{eq:lossrl}
\end{equation}
where $ \reward( \bX ) $ refers to the reward function
%, $\data$ refers to the dataset, 
and $\ent(p)$ denotes the entropy of a distribution $p$ (\ie $ \ent(p(\bX)) = \int p(\bX) \log p(\bX) \text{d} \bX$). 
The last two terms are constants since $\E_{ p_\theta(\bX_{0} ) } $ and $ p_\theta(\bX_{0} ) $ are constants for a fixed dataset. \emph{We drop them in practice.}

Although Eq.~\eqref{eq:lossrl} seems like a natural formulation, direct policy optimization faces significant challenges. First, diffusion models require hundreds of iterative refinements to generate high-quality samples. However, estimating gradients by directly sampling such long decision trajectories from the policy distribution is costly, non-station, and may suffer large variances.
Moreover, for some non-smooth reward functions in a high-dimensional output space, it is difficult for learners to find any highly rewarded samples, especially in early diffusion / reverse steps. 

To tackle these challenges, we propose a novel reward-guided framework, which lies upon the typical entropy regularized reward expectation objective (\ie Eq.~\eqref{eq:lossrl}), for diffusion model training. Particularly, we estimate the gradient of the objective (\ie Eq.~\eqref{eq:lossrl}) w.r.t. policy parameters by sampling from \ul{an exponential reward-aware payoff distribution rather than the policy itself.} Our analysis suggests that by introducing the reward-aware payoff distribution, the entropy regularized reward expectation objective is naturally transformed into \emph{reward re-weighted biased noise prediction loss} for diffusion models.

\subsection{Reward Guided Sampling}

As mentioned above, optimizing $\loss_{RL}$ using SGD is challenging. Motivated by~\cite{norouzi2016reward}, we introduce a {\em exponential payoff distribution} $q_{pd}$ which links Eq.~\eqref{eq:qx_t-1|x,t_x*} and RL objectives:
\begin{equation}
\begin{split}
\bX_{t-1} \sim\; & q_{pd} (\bX_{t-1} \mid \bX_{t}, \bX_{0}) = \int  p( \bX_{t-1} \mid \tilde{\bX}_{t-1} ) q(\tilde{\bX}_{t-1} \mid \bX_{t}, \bX_{0}) \text{d} \tilde{\bX}_{t-1} ,
\end{split}
\label{eq:payoff}
\end{equation}
where:
\begin{equation}
\begin{split}
& \tilde{\bX}_{t-1} \sim q(\tilde{\bX}_{t-1} \mid \bX_{t}, \bX_{0}) \\
&  p( \bX_{t-1} \mid \tilde{\bX}_{t-1} ) = \frac{1}{ Z } [ \exp (\reward ( \bX_{t-1} ) ) \cdot \mathcal{U} ( \tilde{\bX}_{t-1}, d ) ] \\
 & \mathcal{U} ( \tilde{\bX}_{t-1}, d ) =  
\begin{cases}
\frac{1}{\pi d} & \text{if $\| \bX - \bX_{t-1} \|^2 < d $,} \\
0 & \text{otherwise.}
\end{cases}
\end{split}
\label{eq:qpd}
\end{equation}
Here, $Z = \int p( \bX_{t-1} \mid \tilde{\bX}_{t-1} ) \text{d}\bX_{t-1} $ is a constant. Distribution $q$ is identical to Eq.~\eqref{eq:qx_t-1|x,t_x*}. 
$\mathcal{U} ( \tilde{\bX}_{t-1}, d )$ is a `restriction' distribution to force $\tilde{\bX}_{t-1}$ to be adjacent to $\bX_{t-1}$. Intuitively, in diffusion dynamics, intermediate results should move smoothly towards the desired shape (with high reward). The distance between $\bX_{t-1}$ and $\tilde{\bX}_{t-1}$ should be small. Given such a  distribution, we can sample $\bX_{t-1}$ by simply searching $\tilde{\bX}_{t-1}$'s adjacent region for $\bX_{t-1}$s with highest rewards in samples. Concretely,
\begin{equation}
\bX_{t-1} = \tilde{\bX}_{t-1} + \beps_*(\reward(\bX_{t-1})),
\end{equation}
where $\beps_*(\reward(\bX_{t-1}))$ denotes the feasible shift on $\tilde{\bX}_{t-1}$ to maximize $\reward(\bX_{t-1})$. 

Note that \ul{we do not make an assumption on the differentiability} of $\reward(\bX_{t-1})$. Essentially, $\reward(\bX_{t-1})$ is treated as a blackbox.
When $\reward(\bX_{t-1})$ is differentiable, we  solve for $\beps_*(\reward(\bX_{t-1}))$ via gradient-based approaches. Otherwise, one may use non-gradient optimization methods like simulated annealing to obtain $\beps_*(\reward(\bX_{t-1}))$.

% We illustrate such a nested distribution via a probabilistic graphical model (PGM) in Figure~\ref{fig:pgm}.

\subsection{Reward Guided Loss}

One can verify that the global minimum of $\loss_{RL}$, \ie the optimal regularized expected reward, is achieved when the model distribution $ p_\theta $ perfectly matches the exponentiated payoff distribution $ q_{pd} $.
To see this, we re-express the objective function in Eq.~\eqref{eq:lossrl} in terms of a KL divergence between $ p_\theta( \bX_{t-1} \mid \bX_{t} ) $ and $ q_{pd}( \bX_{t-1} \mid \bX_t, \bX_0 ) ) $: 
\begin{equation}
\begin{split}
\loss_{RL} \propto \E_{ \data } 
\sum_{t=1}^{T}  \kl( p_\theta( \bX_{t-1} \mid \bX_{t} ) \mid\mid  q_{pd}( \bX_{t-1} \mid \bX_t, \bX_0 ) ) .
\end{split}
\label{eq:RL-kl} 
\end{equation}
Due to space limitation, we postpone the detailed derivation from Eq.~\eqref{eq:lossrl} Eq.~\eqref{eq:RL-kl} to Appendix~\ref{sec:eq7to11}.

Since the global minimum of Eq.~\eqref{eq:RL-kl} is achieved when $ p_\theta( \bX_{t-1} \mid \bX_{t} ) $ matches $ q_{pd}( \bX_{t-1} \mid \bX_t ) $, we can swap $ p_\theta( \bX_{t-1} \mid \bX_{t} ) $ and $ q_{pd}( \bX_{t-1} \mid \bX_t ) $ in Eq.~\eqref{eq:RL-kl}, without impacting the learning objective: 
\begin{equation}
\begin{split}
\loss_{RMLE} \propto  \E_{ \data }
\sum_{t=1}^{T} \kl( q_{pd}( \bX_{t-1} \mid \bX_t, \bX_0 )  \mid\mid  p_\theta( \bX_{t-1} \mid \bX_{t}) ) .
% \\ = & - \E_{ \data } 
%\sum_{t=1}^T  \left[ 
%\E_{ q_{pd}( \bX_{t-1} \mid \bX_t, \bX_0 ) } \log p_\theta(\bX_{t-1} \mid \bX_{t} )   \right]  \\
%& + \text{Constant}
\end{split}
\label{eq:RMLE-kl} 
\end{equation}

The objective functions $\loss_{RL}$ and $\loss_{RMLE}$, have the same global optimum of $p$, but they optimize a KL divergence in opposite directions. 
When optimizing 
$\loss_{RMLE}$, one can draw unbiased samples from the stationary exponential payoff distribution $ q_{pd} $ instead of the model $p$ itself. 

From Eq.~\eqref{eq:RMLE-kl}, we derive the concrete learning objective:
\begin{equation}
\begin{split}
\min_\theta \; & \E_{ \data }
\bigg \{
\sum_{t=1}^{T}
\int q_{pd} (\log q_{pd} - \log p_\theta ) \text{d} \bX_{t-1} \bigg \} \\
\propto & \E_{ \data }
\bigg \{ 
\sum_{t=1}^{T} 
 \E_{\bX_t} \exp(  r (\bX_{t-1}) ) \frac{ \| \bX_{t-1} - \bmu_\theta(\bX_{t}, t) \|^2 }{2\beta_t} \bigg \},  \\
% \propto \; & \E_{ \data }
% \bigg \{ 
% \sum_{t=1}^{T} 
% - \E_{\bX_t} r (\bX_{t-1}) \frac{ \| \bX_{t-1} - \bX_{t} \|^2}{2\beta_t} \bigg \} ,
% & \quad \| \bX_{t-1}  -  \bX_t \|^2 \leq \tau,
\end{split}
\label{eq:RMLE-kl-obj}
\end{equation}
where we use $q_{pd}$ and $p$ to denote $q_{pd}( \bX_{t-1} \mid \bX_t, \bX_0 )$ and $p_\theta( \bX_{t-1} \mid \bX_{t} )$ for abbreviation.  Please refer to Appendix for the detailed derivations.

To implement Eq.~\eqref{eq:RMLE-kl-obj},, we reformulate $\bX_{t-1}$ and $\bmu_\theta(\bX_t,t)$ in Eq.~\eqref{eq:RMLE-kl-obj}  following Sec.~\ref{sec:background} and Eq.~\eqref{eq:payoff} as: 
\begin{equation}
\begin{split}
& \bX_{t} =  \sqrt{ \bar{\alpha}_t }  \bX_0 + \sqrt{1 - \bar{\alpha}_t } \beps \quad\quad
\tilde{\bX}_{t-1} =  \frac{ \sqrt{ \bar{\alpha}_{t-1} } \beta_t }{ 1 - \bar{\alpha}_t } \bX_0 + \frac{ \sqrt{\alpha_t}(1 - \bar{\alpha}_{t - 1} ) }{ 1 - \bar{\alpha}_t } \bX_t \\
& \bX_{t-1} \sim q_{pd}( \bX_{t-1} \mid \bX_t, \bX_0 ) = \tilde{\bX}_{t-1} + \beps_*(\reward(\bX_{t-1})) \\
& \bmu_\theta(\bX_t, t) =  \frac{1}{ \sqrt{ \alpha}_t } \bX_{t} - \frac{\beta_t \beps_\theta (\bX_{t}, t) }{ \sqrt{ \alpha_t (1 - \bar{\alpha}_t ) } },
\label{eq:Xt-1}
\end{split}
\end{equation}
where $\beps_*(\reward(\bX_{t-1}))$ denotes the shifting vector suggested by the pay-off distribution. 
% \emph{In practices, one can manually manipulate $\beps_*(\reward(\bX_{t-1}))$ to get $\bX_{t-1}$ with different $\reward(\bX_{t-1})$. This process is equivalent to drawing samples from $q_{pd}$}. 

To this point, we have both $ \bX_{t-1}$ and $ \mu_\theta(\bX_t, t) $ estimated. Thus, we can rewrite Eq.~\eqref{eq:RMLE-kl-obj} as a compact formulation:
\begin{equation}
\min_\theta \; \E_{ \data } \Bigg\{
\sum_{t=1}^{T} 
- \E_{\bX_t} \frac{ \reward(\bX_{t-1})}{ 2\beta_t } \Bigg\| \beps_*(\reward(\bX_{t-1})) + \frac{\beta_t}{\sqrt{ \alpha_t (1 - \bar{\alpha}_t )}}\left(\beps -  \beps_\theta (\bX_{t}, t)\right)   \Bigg\|^2   \Bigg\}. 
\label{eq:RMLE-kl-obj-2} 
\end{equation}
Such a formulation is obtained by simply substitute $ \bX_{t-1}$ and $ \mu_\theta(\bX_t, t) $ in Eq.~\eqref{eq:RMLE-kl-obj} according to Eq.~\eqref{eq:Xt-1}.

\textbf{Remark} Compared to the learning objective of the vanilla DDPM (Eq.~\eqref{eq:loss_ll}), the reward-guided diffusion model (Eq.~\eqref{eq:RMLE-kl-obj} or \eqref{eq:RMLE-kl-obj-2}) suggests a \emph{reward re-weighted biased noise prediction loss}, which favors (1) the samples with high rewards and (2) critic changes in the reversed process of diffusion models that lead to high rewards.
Besides, unlike conventional RL objective (Eq.~\eqref{eq:lossrl}), in Eq.~\eqref{eq:RMLE-kl-obj-2} all the sampled terms (\ie $\beps$, $\beps_*(\reward(\bX_{t-1}))$ and $\bX_{t-1}$) are drawn from stationary distribution (Gaussian or the proposed payoff distribution) instead of the intermediate policy. 
Such a sampling strategy reduces the variances of estimated policy gradients, and pilots the reversed process of diffusion models toward more highly-rewarded directions.

\subsection{Training}

Since the reward is often sparse in a high-dimensional output space, `smart' model initialization (pre-training) instead of random initialization is needed. That is to say, at the beginning of the training, we use the maximum likelihood estimation (MLE) to \ul{pre-train the diffusion model on the training set $\data$.} Then we use the proposed \modelname to adjust the diffusion model according to the guidance of the reward function.

We summarize the learning algorithm in Alg.~\ref{alg_t}. 
The sampling process of the proposed \modelname is identical to vanilla diffusion models. 
% and Alg.~\ref{alg_s}, respectively.

\begin{algorithm}
    \renewcommand{\algorithmicrequire}{\textbf{Input:}}
	\renewcommand{\algorithmicensure}{\textbf{Output:}}
	\caption{RLM Training Process}
	\label{alg_t}
	\begin{algorithmic}[1]
	\REQUIRE Sample $\bX_0$, equivariant encoder $\beps_{\theta}$
    \STATE Pre-train the diffusion model by optimizing Eq.~\eqref{eq:loss}.
    \REPEAT
    \STATE Sample $t \sim \mathcal{U}(0 \cdots T)$.
    \STATE Calculate $\bX_t$, $\bX_{t-1}$ and $\bmu_\theta(\bX_t, t)$ according to Eq.~\eqref{eq:Xt-1}. 
    \STATE Take a gradient descent step on Eq.~\eqref{eq:RMLE-kl-obj-2}.
    \UNTIL Converged 
	\end{algorithmic}  
\end{algorithm}

\section{Experiments}

In this section, we report the experimental results on multiple benchmark data sets crossing two tasks (i.e., 3D shape generation and molecule generation). The results show that the proposed \modelname outperforms state-of-the-art diffusion models in controlled generation.

\subsection{Controlled 3D Shape Generation}

% \subsubsection{Experiment Settings}\label{sec:setup1}

\paragraph{Dataset} To evaluate the performance of controllable 3D shape  generation, we adopt a fine-grained 3D shape dataset named FG3D~\cite{liu2021fine}, which is built upon the well-known 3D shape datasets such as ShapeNet~\cite{chang2015shapenet}, Yobi3D~\cite{yobi3d} and 3D Warehouse~\cite{goldfeder2008autotagging}. FG3D contains 25,552 shapes from three general categories including Airplane, Car and Chair, which are further labeled into 13, 20 and 33 sub-categories. For each shape, we randomly sample 2048 points via Open3D\footnote{www.open3d.org} to obtain the point clouds and normalize the point clouds to zero mean and unit variance. Finally, we randomly split each sub-category into training, validation and testing sets by the ratio 90\%, 5\% and 5\%, respectively.

\paragraph{Evaluation Metrics} The generation task in this section is \ul{to force the generator to generate samples of a specific sub-category.} We expect the generation samples \ul{to be as similar to real samples from the targeted sub-category as possible.} Here, we choose sub-categories \ul{`helicopter', `bus' and `bar (chair)'} from general categories Airplane, Car and Chair, respectively.

We adopt three commonly used metrics in existing 3D shape generation literature~\cite{shu20193d,tang2022warpinggan,cai2020learning}, \ie \textit{Minimum Matching Distance (MMD)}: (the averaged distance between generated shapes and shapes in the reference set); \textit{Coverage (COV)}: (the fraction of shapes that can be matched as a nearest shape with the generated shape); \textit{Jenson-Shannon divergence (JSD)}: (the distance between the generated set and the groundtruth).
\ul{Good methods should have a low MMD, high COV and low JSD.} 

For distance evaluation metrics in MMD and COV, we adopt both Chamfer distance (CD) and Earth Mover distance (EMD).
Besides, following existing papers~\cite{yang2019pointflow,luo2021diffusion,cai2020learning}, we normalize both generated point clouds and groundtruth references into a bounding box of [-1, 1] to let the evaluation focus on shape rather than scale. 

\paragraph{Baselines} We evaluate the shape generation capability of our method against two representative state-of-the-art controllable diffusion models, including 3D-DDPM~\cite{luo2021diffusion} (\textit{introduce categorical conditional variable for controlled generation}) and Test-Phase Manipulation (TPM)~\cite{dhariwal2021diffusion} (\textit{manipulate the sampling phase via a pre-trained classifier/regressor}). The descriptions of these baselines are already detailed in Sec.~\ref{sec:related}. 
\ul{For both the baselines and the proposed method, we use the neural network architecture in}~\citet{luo2021diffusion} \ul{as backbone of the diffusion model.} The parameters are set as the original papers suggest. 

\paragraph{Implementation Details} 
For all the experiments in this section, we set the number of steps, \ie $T$, in the diffusion process to 100. The noise scheduling factor of step $t$, \ie $\beta_t$, is set to linearly increase  from $\beta_1 = 0.0001$ to $\beta_T = 0.05$. We choose to implement the noise estimator $\beps_\theta$ as a 7-layer MLP with concatsquash~\cite{grathwohl2018ffjord} layers and LeakyReLU. The dimension of concatsquash layers are (3-128-256-512-256-128-3). 
The \emph{rewards} in \modelname come from PointNet~\cite{qi2017pointnet} classifiers.

\begin{table*}[t!]
\centering
\caption{Performance on the FG3D dataset. The classes in parentheses indicate the sub-category that we force the models to generate.  CD, and EMD distances are multiplied by $10^1$.}
\label{table:results1}
\begin{tabular}{cc|cc|cc|c}
\toprule
                          &         & \multicolumn{2}{c}{MMD $\downarrow$ } & \multicolumn{2}{|c|}{COV (\% $\uparrow$)}   & \multicolumn{1}{c}{JSD $\downarrow$ } \\
Shape                     & Model   & CD         & EMD        & CD  & EMD &  -   \\
\midrule
\multirow{3}{*}{Airplane (helicopter)} & 3D-DDPM &    0.176        &  1.642          &   34.29  &  42.86   &  0.2321   \\
                          & TPM     &    8.386        &  10.783          &   8.57  &  8.57   &  0.9988   \\
                          & \modelname  &    \textbf{0.113}        &    \textbf{1.340}        &  \textbf{51.43}   &  \textbf{42.86}   &  \textbf{0.1496}   \\
\midrule
\multirow{3}{*}{Car (bus)}      & 3D-DDPM &   0.101         &    1.150        &  16.00   &  11.50   &  0.2326   \\
                          & TPM     &    9.022        &     6.772       &  2.00   &  2.00   &  0.9976   \\
                          & \modelname  &    \textbf{0.068}        &     \textbf{0.938}       &  \textbf{48.00}   &  \textbf{40.00}   &   \textbf{0.0979}  \\
\midrule
\multirow{3}{*}{Chair (bar)}   & 3D-DDPM &   0.474         &    2.682        &  31.11   &  42.22   &  0.1624   \\
                          & TPM     &    5.143        &     1.226       &  8.89   &  2.22   &  0.9390   \\
                          & \modelname  &    \textbf{0.253}        &     \textbf{2.091}       &  \textbf{57.78}   &  \textbf{57.78}   &   \textbf{0.1063}  \\
\bottomrule
\end{tabular}
\end{table*}

\subsubsection{Results Analysis}\label{sec:results1}

\paragraph{Quantitative evaluation}
Table~\ref{table:results1} reports the quantitative results of the proposed method and all the baselines. From the results, we can see that the proposed \modelname consistently outperforms all the baselines for all the metrics by significant margins. Particularly, low MMDs suggest that the proposed \modelname enable generated shapes to be similar to the shapes within the desired sub-categories in terms of both the spatial and feature space. The high COVs suggest that our generated shapes have a good coverage of the shapes in the desired sub-categories. 

\begin{wrapfigure}{r}{0.45\textwidth}
\vspace{-5mm}
\begin{center}
      \includegraphics[width=\linewidth]{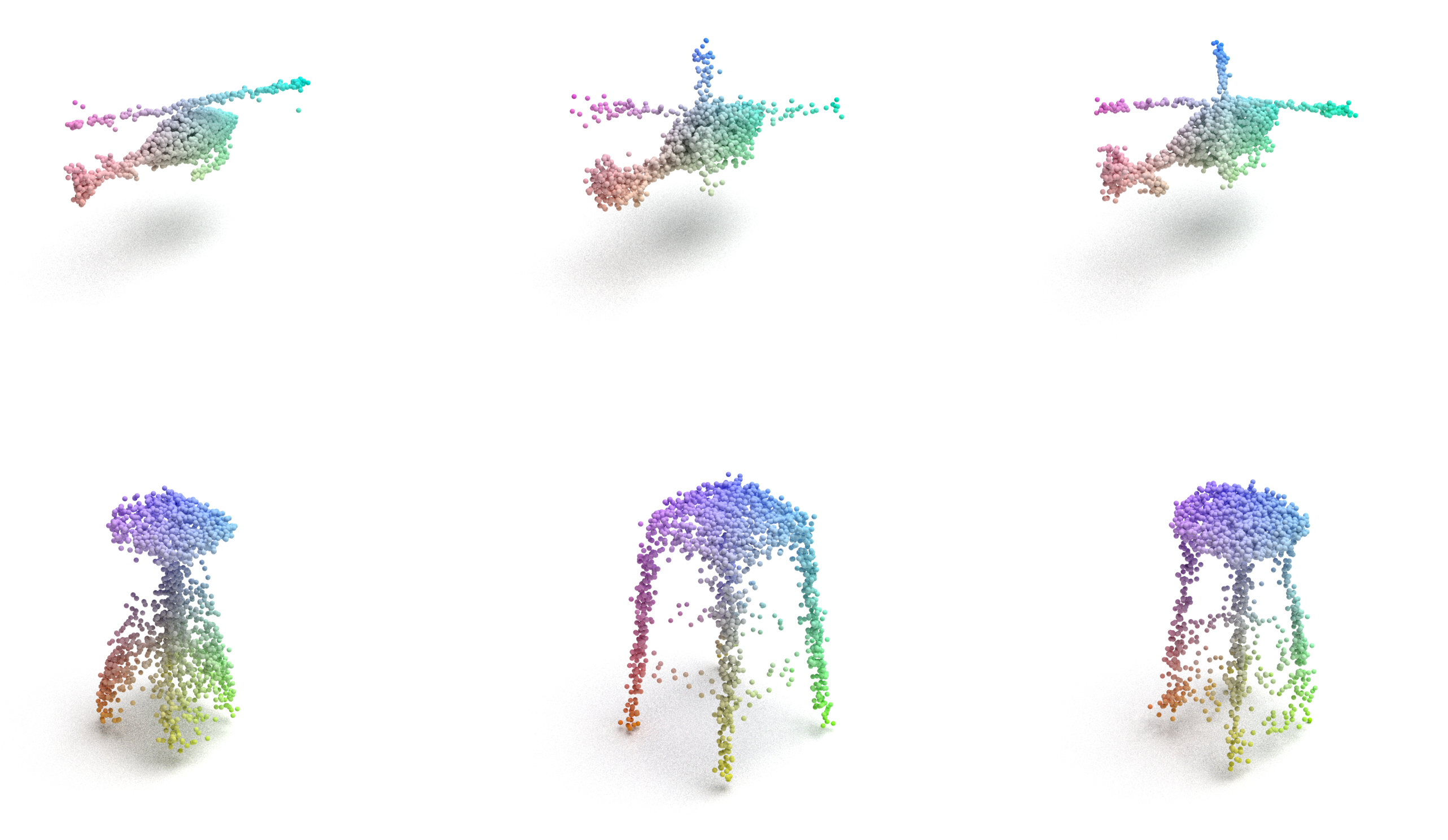}
\end{center}
\caption{Generated point clouds with desired sub-categories, i.e., helicopter (airplane) and bar (chair).}
\label{fig:pcd}
\end{wrapfigure}

By outperforming existing controllable diffusion model training frameworks, the proposed \modelname demonstrates its advantage in guiding diffusion models towards the desired sub-categories. As a comparison, gradients from pre-trained classifiers, which are used in the test-phase controlling method (i.e., TPM~\cite{dhariwal2021diffusion}), fail to provide enough guidance toward the desired sub-categories. Besides, despite outperforming TPM~\cite{dhariwal2021diffusion}, contextual representations in (conditional) 3D-DDPM~\cite{luo2021diffusion} cannot match the performance of \modelname. \textcolor{black}{This is because (conditional) 3D-DDPM~\cite{luo2021diffusion} treats conditional information and general shape structure information as disentangled, via separated representations}. However, since shapes from different sub-categories still share common characteristics, the learned contextual representations may not offer `clean' enough guidance toward generating shapes of a specific sub-category. \textcolor{black}{On the contrary, the proposed reward-guided objective in \modelname effectively forces the reversed  trajectories toward the high-reward directions.} That is why \modelname consistently outperforms 3D-DDPM~\cite{luo2021diffusion}.

\paragraph{Visualization} 
Figure~\ref{fig:pcd} shows some visual comparison results. 
Here, we randomly pick several shape generated by \modelname given specific sub-categories.
From these results, we can see that the point clouds generated by our method consists of fine details (e.g., propeller blade of helicopters) and little noise and few outlier points. The points in the shapes are also distributed uniformly, with no significant ``holes''.

\subsection{Controlled 3D Molecular Generations}

% \begin{figure*}
%     \centering
%     \includegraphics[height=3cm]{example-image-a}
%     \caption{Examples of generated molecules.}
%     \label{fig:my_label}
% \end{figure*}

% \subsubsection{Experiment Settings}

\paragraph{Dataset} QM9 \citep{ramakrishnan2014quantum} is a dataset of 130k stable and synthetically accessible organic molecules with up to 9 heavy atoms (29 atoms including hydrogens). In this section, we train diffusion models to generate atoms' (1) 3-dimensional coordinates; (2) types (H, C, N, O, F) and (3) integer-valued atom charges. 
We use the train/val/test partitions introduced in \citet{anderson2019cormorant} (train/val/test: 100K/18K/13K samples) for evaluation.

\paragraph{Evaluation Metrics} Our goal here is to generate molecules targeting some desired properties while at the same time not harming general generation quality (e.g., molecules' validity (the proportion of atoms with right valency) and stability, etc.). In such a scenario, a molecule is represented as a point cloud, in which each point denotes a single atom and has its own (atom) type.
Following \cite{satorras2021en_flows}, for each pair of atoms,
we use the distance between them and the atoms' types to predict bonds (single, double, triple, or none) between atoms. 

In this section, we consider optimizing two desired properties: (1) quantitative estimate of drug-likeness (QED)~\cite{bickerton2012quantifying} (how likely a molecule is a potential drug candidate based on marketed drug molecules) and (2) synthetic accessibility score (SA) (the difficulty of drug synthesis), which are crucial in drug discovery domain. \ul{A good method should have a high averaged QED and SA. Note that we conduct separated experiments for these two properties, which means only one property is considered in a single experiment.}
We adopt widely-used open-source cheminformatics software RDKit\footnote{https://www.rdkit.org} to calculate the properties above.

\paragraph{Baselines} In this section, we adopt E(3) equivariant diffusion model (EDM)~\cite{hoogeboom2022equivariant}, which achieves SOTA performance in 3D conditional molecule generation, as the baseline method. EDM specifies a diffusion process that operates on both continuous coordinates and categorical atom types. Particularly, EDM adopts equivariant graph neural networks (EGNN)~\cite{satorras2021egnn} to model molecular geometries, which are equivariant to the action of rotations, reflections and translations. \ul{For both the baseline and the proposed method, we use the neural network architecture specified in}~\citet{hoogeboom2022equivariant} \ul{as the backbone.} We exclude TPM~\cite{dhariwal2021diffusion} as a comparison method because it requires gradients from a pre-trained classifier (scoring function) to guide the controlled generation. Nevertheless, the property calculators for QED and SA are non-differentiable.

\paragraph{Implementation Details} 
For all the experiments in this section, we set the number of steps $T$ in the diffusion process to 1,000. \textcolor{black}{The noise scheduling factor of step $t$, \ie $\beta_t$, is set to the cosine noise
schedule introduced in~\citet{nichol2021improved,hoogeboom2022equivariant}. }
We choose to implement the noise estimator as EGNNs with 256 hidden features and 9 layers.
The \emph{rewards} in \modelname come from RDKit QED and SA calculator.

\subsubsection{Result Analysis}

The results of controlled molecule generation are reported in Table \ref{tab:qm9_results}. Compared with EDM, the proposed reward-guided framework \modelname generates molecules with higher QED and SA scores without hurting their general chemistry characteristics (\ie high rate of validity and stability). Such results again demonstrate the superiority of reward-guided diffusion models over existing conditional diffusion models, such as EDM.

\begin{table}[h!]
\centering
\caption{Molecule stability (Mol Stable), Validity (Valid),  QED and SA across 3 runs on QM9, each drawing 1,000 samples from the model.}
\begin{tabular}{cc|cc}
\toprule
\multicolumn{2}{c|}{Methods}                            & EDM   & \modelname \\
\midrule
\multicolumn{1}{c|}{\multirow{2}{*}{\begin{tabular}[c]{@{}c@{}}General\\  Properties\end{tabular}} }       & Mol Stable (\%) & 90.7  & 90.5    \\
\multicolumn{1}{c|}{}                                         & Valid (\%)      & 91.2  & 91.4    \\
\midrule
\multicolumn{1}{c|}{\multirow{2}{*}{\begin{tabular}[c]{@{}c@{}}Properties \\ to Optimize\end{tabular}} } & Avg. QED        & 0.461 & \textbf{0.542}   \\
\multicolumn{1}{c|}{}                                         & Avg. SA       & 4.41 & \textbf{5.87}   \\
\bottomrule
\end{tabular}
\label{tab:qm9_results}
\end{table}

Moreover, it is worth emphasising that the molecular point clouds are much more sparse and discontinuous than 3D shape clouds.
This characteristics makes it difficult for conventional RL learners to find highly-rewarded refinements in reversed trajectories. Nevertheless, by taking advantage of the reward-aware payoff distribution, the proposed \modelname successfully overcomes such difficulty and achieves impressive performance.

\section{Conclusions}

This paper presents a reward-guided learning framework for diffusion models that enables flexible controlled generation. Unlike the conventional RL framework, the proposed \modelname estimates policy (model) gradients through samples from the stationary reward-aware payoff distribution rather than the policy itself. Thus, the learning objective is turned into a \emph{reward re-weighted biased noise prediction loss}, which can effectively guide the reversed process towards highly-rewarded directions and simultaneously reduce variances of the estimated gradients. Experimental results on 3D shape generation and molecular generation tasks show that the proposed framework outperforms existing controlled diffusion models by a clear margin. In the future, we plan to explore more effective sampling strategies for payoff distributions to improve the model performance.

\bibliographystyle{plainnat}
\bibliography{egbib}

\begin{thebibliography}{45}
\providecommand{\natexlab}[1]{#1}
\providecommand{\url}[1]{\texttt{#1}}
\expandafter\ifx\csname urlstyle\endcsname\relax
  \providecommand{\doi}[1]{doi: #1}\else
  \providecommand{\doi}{doi: \begingroup \urlstyle{rm}\Url}\fi

\bibitem[Anderson et~al.(2019)Anderson, Hy, and Kondor]{anderson2019cormorant}
Brandon Anderson, Truong~Son Hy, and Risi Kondor.
\newblock Cormorant: Covariant molecular neural networks.
\newblock \emph{Advances in neural information processing systems}, 32, 2019.

\bibitem[Bickerton et~al.(2012)Bickerton, Paolini, Besnard, Muresan, and
  Hopkins]{bickerton2012quantifying}
G~Richard Bickerton, Gaia~V Paolini, J{\'e}r{\'e}my Besnard, Sorel Muresan, and
  Andrew~L Hopkins.
\newblock Quantifying the chemical beauty of drugs.
\newblock \emph{Nature chemistry}, 4\penalty0 (2):\penalty0 90--98, 2012.

\bibitem[Cai et~al.(2020)Cai, Yang, Averbuch-Elor, Hao, Belongie, Snavely, and
  Hariharan]{cai2020learning}
Ruojin Cai, Guandao Yang, Hadar Averbuch-Elor, Zekun Hao, Serge Belongie, Noah
  Snavely, and Bharath Hariharan.
\newblock Learning gradient fields for shape generation.
\newblock In \emph{European Conference on Computer Vision}, pages 364--381.
  Springer, 2020.

\bibitem[Chang et~al.(2015)Chang, Funkhouser, Guibas, Hanrahan, Huang, Li,
  Savarese, Savva, Song, Su, et~al.]{chang2015shapenet}
Angel~X Chang, Thomas Funkhouser, Leonidas Guibas, Pat Hanrahan, Qixing Huang,
  Zimo Li, Silvio Savarese, Manolis Savva, Shuran Song, Hao Su, et~al.
\newblock Shapenet: An information-rich 3d model repository.
\newblock \emph{arXiv preprint arXiv:1512.03012}, 2015.

\bibitem[Chen et~al.(2020)Chen, Zhang, Zen, Weiss, Norouzi, and
  Chan]{chen2020wavegrad}
Nanxin Chen, Yu~Zhang, Heiga Zen, Ron~J Weiss, Mohammad Norouzi, and William
  Chan.
\newblock Wavegrad: Estimating gradients for waveform generation.
\newblock \emph{arXiv preprint arXiv:2009.00713}, 2020.

\bibitem[Choi et~al.(2021)Choi, Kim, Jeong, Gwon, and Yoon]{choi2021ilvr}
Jooyoung Choi, Sungwon Kim, Yonghyun Jeong, Youngjune Gwon, and Sungroh Yoon.
\newblock Ilvr: Conditioning method for denoising diffusion probabilistic
  models.
\newblock \emph{arXiv preprint arXiv:2108.02938}, 2021.

\bibitem[Dhariwal and Nichol(2021)]{dhariwal2021diffusion}
Prafulla Dhariwal and Alexander Nichol.
\newblock Diffusion models beat gans on image synthesis.
\newblock \emph{Advances in Neural Information Processing Systems},
  34:\penalty0 8780--8794, 2021.

\bibitem[Goldfeder and Allen(2008)]{goldfeder2008autotagging}
Corey Goldfeder and Peter Allen.
\newblock Autotagging to improve text search for 3d models.
\newblock In \emph{Proceedings of the 8th ACM/IEEE-CS joint conference on
  Digital libraries}, pages 355--358, 2008.

\bibitem[Grathwohl et~al.(2018)Grathwohl, Chen, Bettencourt, Sutskever, and
  Duvenaud]{grathwohl2018ffjord}
Will Grathwohl, Ricky~TQ Chen, Jesse Bettencourt, Ilya Sutskever, and David
  Duvenaud.
\newblock Ffjord: Free-form continuous dynamics for scalable reversible
  generative models.
\newblock \emph{arXiv preprint arXiv:1810.01367}, 2018.

\bibitem[Haarnoja(2018)]{haarnoja2018acquiring}
Tuomas Haarnoja.
\newblock \emph{Acquiring diverse robot skills via maximum entropy deep
  reinforcement learning}.
\newblock University of California, Berkeley, 2018.

\bibitem[Ho and Salimans(2022)]{ho2022classifier}
Jonathan Ho and Tim Salimans.
\newblock Classifier-free diffusion guidance.
\newblock \emph{arXiv preprint arXiv:2207.12598}, 2022.

\bibitem[Ho et~al.(2020)Ho, Jain, and Abbeel]{ho2020denoising}
Jonathan Ho, Ajay Jain, and Pieter Abbeel.
\newblock Denoising diffusion probabilistic models.
\newblock \emph{arXiv preprint arXiv:2006.11239}, 2020.

\bibitem[Ho et~al.(2022)Ho, Saharia, Chan, Fleet, Norouzi, and
  Salimans]{ho2022cascaded}
Jonathan Ho, Chitwan Saharia, William Chan, David~J Fleet, Mohammad Norouzi,
  and Tim Salimans.
\newblock Cascaded diffusion models for high fidelity image generation.
\newblock \emph{J. Mach. Learn. Res.}, 23:\penalty0 47--1, 2022.

\bibitem[Hoogeboom et~al.(2022)Hoogeboom, Satorras, Vignac, and
  Welling]{hoogeboom2022equivariant}
Emiel Hoogeboom, V{\i}ctor~Garcia Satorras, Cl{\'e}ment Vignac, and Max
  Welling.
\newblock Equivariant diffusion for molecule generation in 3d.
\newblock In \emph{International Conference on Machine Learning}, pages
  8867--8887. PMLR, 2022.

\bibitem[https://www.yobi3d.com()]{yobi3d}
https://www.yobi3d.com.
\newblock Yobi3d-free 3d model search engine [online].

\bibitem[Kappen et~al.(2012)Kappen, G{\'o}mez, and Opper]{kappen2012optimal}
Hilbert~J Kappen, Vicen{\c{c}} G{\'o}mez, and Manfred Opper.
\newblock Optimal control as a graphical model inference problem.
\newblock \emph{Machine learning}, 87\penalty0 (2):\penalty0 159--182, 2012.

\bibitem[Kingma et~al.(2021)Kingma, Salimans, Poole, and
  Ho]{kingma2021variational}
Diederik~P Kingma, Tim Salimans, Ben Poole, and Jonathan Ho.
\newblock Variational diffusion models.
\newblock \emph{arXiv preprint arXiv:2107.00630}, 2, 2021.

\bibitem[Klokov et~al.(2020)Klokov, Boyer, and Verbeek]{klokov2020discrete}
Roman Klokov, Edmond Boyer, and Jakob Verbeek.
\newblock Discrete point flow networks for efficient point cloud generation.
\newblock In \emph{European Conference on Computer Vision}, pages 694--710.
  Springer, 2020.

\bibitem[Kong et~al.(2020)Kong, Ping, Huang, Zhao, and
  Catanzaro]{kong2020diffwave}
Zhifeng Kong, Wei Ping, Jiaji Huang, Kexin Zhao, and Bryan Catanzaro.
\newblock Diffwave: A versatile diffusion model for audio synthesis.
\newblock \emph{arXiv preprint arXiv:2009.09761}, 2020.

\bibitem[Li et~al.(2021)Li, Li, Hui, and Fu]{li2021sp}
Ruihui Li, Xianzhi Li, Ka-Hei Hui, and Chi-Wing Fu.
\newblock Sp-gan: Sphere-guided 3d shape generation and manipulation.
\newblock \emph{ACM Transactions on Graphics (TOG)}, 40\penalty0 (4):\penalty0
  1--12, 2021.

\bibitem[Liu et~al.(2021)Liu, Han, Liu, and Zwicker]{liu2021fine}
Xinhai Liu, Zhizhong Han, Yu-Shen Liu, and Matthias Zwicker.
\newblock Fine-grained 3d shape classification with hierarchical part-view
  attention.
\newblock \emph{IEEE Transactions on Image Processing}, 30:\penalty0
  1744--1758, 2021.

\bibitem[Luo and Hu(2021)]{luo2021diffusion}
Shitong Luo and Wei Hu.
\newblock Diffusion probabilistic models for 3d point cloud generation.
\newblock In \emph{Proceedings of the IEEE/CVF Conference on Computer Vision
  and Pattern Recognition}, pages 2837--2845, 2021.

\bibitem[Mittal et~al.(2022)Mittal, Cheng, Singh, and
  Tulsiani]{mittal2022autosdf}
Paritosh Mittal, Yen-Chi Cheng, Maneesh Singh, and Shubham Tulsiani.
\newblock Autosdf: Shape priors for 3d completion, reconstruction and
  generation.
\newblock In \emph{Proceedings of the IEEE/CVF Conference on Computer Vision
  and Pattern Recognition}, pages 306--315, 2022.

\bibitem[Mnih et~al.(2013)Mnih, Kavukcuoglu, Silver, Graves, Antonoglou,
  Wierstra, and Riedmiller]{mnih2013playing}
Volodymyr Mnih, Koray Kavukcuoglu, David Silver, Alex Graves, Ioannis
  Antonoglou, Daan Wierstra, and Martin Riedmiller.
\newblock Playing atari with deep reinforcement learning.
\newblock \emph{arXiv preprint arXiv:1312.5602}, 2013.

\bibitem[Mnih et~al.(2016)Mnih, Badia, Mirza, Graves, Lillicrap, Harley,
  Silver, and Kavukcuoglu]{mnih2016asynchronous}
Volodymyr Mnih, Adria~Puigdomenech Badia, Mehdi Mirza, Alex Graves, Timothy
  Lillicrap, Tim Harley, David Silver, and Koray Kavukcuoglu.
\newblock Asynchronous methods for deep reinforcement learning.
\newblock In \emph{International conference on machine learning}, pages
  1928--1937. PMLR, 2016.

\bibitem[Nichol and Dhariwal(2021)]{nichol2021improved}
Alex Nichol and Prafulla Dhariwal.
\newblock Improved denoising diffusion probabilistic models.
\newblock \emph{arXiv preprint arXiv:2102.09672}, 2021.

\bibitem[Norouzi et~al.(2016)Norouzi, Bengio, Jaitly, Schuster, Wu, Schuurmans,
  et~al.]{norouzi2016reward}
Mohammad Norouzi, Samy Bengio, Navdeep Jaitly, Mike Schuster, Yonghui Wu, Dale
  Schuurmans, et~al.
\newblock Reward augmented maximum likelihood for neural structured prediction.
\newblock \emph{Advances In Neural Information Processing Systems}, 29, 2016.

\bibitem[Peters and Schaal(2007)]{peters2007reinforcement}
Jan Peters and Stefan Schaal.
\newblock Reinforcement learning by reward-weighted regression for operational
  space control.
\newblock In \emph{Proceedings of the 24th international conference on Machine
  learning}, pages 745--750, 2007.

\bibitem[Qi et~al.(2017)Qi, Su, Mo, and Guibas]{qi2017pointnet}
Charles~R Qi, Hao Su, Kaichun Mo, and Leonidas~J Guibas.
\newblock Pointnet: Deep learning on point sets for 3d classification and
  segmentation.
\newblock In \emph{Proceedings of the IEEE conference on computer vision and
  pattern recognition}, pages 652--660, 2017.

\bibitem[Ramakrishnan et~al.(2014)Ramakrishnan, Dral, Rupp, and
  Von~Lilienfeld]{ramakrishnan2014quantum}
Raghunathan Ramakrishnan, Pavlo~O Dral, Matthias Rupp, and O~Anatole
  Von~Lilienfeld.
\newblock Quantum chemistry structures and properties of 134 kilo molecules.
\newblock \emph{Scientific data}, 1\penalty0 (1):\penalty0 1--7, 2014.

\bibitem[Ramasinghe et~al.(2020)Ramasinghe, Khan, Barnes, and
  Gould]{ramasinghe2020spectral}
Sameera Ramasinghe, Salman Khan, Nick Barnes, and Stephen Gould.
\newblock Spectral-gans for high-resolution 3d point-cloud generation.
\newblock In \emph{2020 IEEE/RSJ International Conference on Intelligent Robots
  and Systems (IROS)}, pages 8169--8176. IEEE, 2020.

\bibitem[Rombach et~al.(2022)Rombach, Blattmann, Lorenz, Esser, and
  Ommer]{rombach2022high}
Robin Rombach, Andreas Blattmann, Dominik Lorenz, Patrick Esser, and Bj{\"o}rn
  Ommer.
\newblock High-resolution image synthesis with latent diffusion models.
\newblock In \emph{Proceedings of the IEEE/CVF Conference on Computer Vision
  and Pattern Recognition}, pages 10684--10695, 2022.

\bibitem[Saharia et~al.(2022{\natexlab{a}})Saharia, Chan, Chang, Lee, Ho,
  Salimans, Fleet, and Norouzi]{saharia2022palette}
Chitwan Saharia, William Chan, Huiwen Chang, Chris Lee, Jonathan Ho, Tim
  Salimans, David Fleet, and Mohammad Norouzi.
\newblock Palette: Image-to-image diffusion models.
\newblock In \emph{ACM SIGGRAPH 2022 Conference Proceedings}, pages 1--10,
  2022{\natexlab{a}}.

\bibitem[Saharia et~al.(2022{\natexlab{b}})Saharia, Ho, Chan, Salimans, Fleet,
  and Norouzi]{saharia2022image}
Chitwan Saharia, Jonathan Ho, William Chan, Tim Salimans, David~J Fleet, and
  Mohammad Norouzi.
\newblock Image super-resolution via iterative refinement.
\newblock \emph{IEEE Transactions on Pattern Analysis and Machine
  Intelligence}, 2022{\natexlab{b}}.

\bibitem[Satorras et~al.(2021{\natexlab{a}})Satorras, Hoogeboom, Fuchs, Posner,
  and Welling]{satorras2021en_flows}
Victor~Garcia Satorras, Emiel Hoogeboom, Fabian Fuchs, Ingmar Posner, and Max
  Welling.
\newblock E(n) equivariant normalizing flows.
\newblock \emph{Advances in Neural Information Processing Systems}, 34,
  2021{\natexlab{a}}.

\bibitem[Satorras et~al.(2021{\natexlab{b}})Satorras, Hoogeboom, and
  Welling]{satorras2021egnn}
Victor~Garcia Satorras, Emiel Hoogeboom, and Max Welling.
\newblock E (n) equivariant graph neural networks.
\newblock \emph{arXiv preprint arXiv:2102.09844}, 2021{\natexlab{b}}.

\bibitem[Shu et~al.(2019)Shu, Park, and Kwon]{shu20193d}
Dong~Wook Shu, Sung~Woo Park, and Junseok Kwon.
\newblock 3d point cloud generative adversarial network based on tree
  structured graph convolutions.
\newblock In \emph{Proceedings of the IEEE/CVF international conference on
  computer vision}, pages 3859--3868, 2019.

\bibitem[Sohl{-}Dickstein et~al.(2015)Sohl{-}Dickstein, Weiss, Maheswaranathan,
  and Ganguli]{sohldickstein2015diffusion}
Jascha Sohl{-}Dickstein, Eric~A. Weiss, Niru Maheswaranathan, and Surya
  Ganguli.
\newblock Deep unsupervised learning using nonequilibrium thermodynamics.
\newblock In Francis~R. Bach and David~M. Blei, editors, \emph{Proceedings of
  the 32nd International Conference on Machine Learning, {ICML}}, 2015.

\bibitem[Song and Ermon(2019)]{song2019estimatinggradients}
Yang Song and Stefano Ermon.
\newblock Generative modeling by estimating gradients of the data distribution.
\newblock \emph{CoRR}, abs/1907.05600, 2019.
\newblock URL \url{http://arxiv.org/abs/1907.05600}.

\bibitem[Song et~al.(2021)Song, Shen, Xing, and Ermon]{song2021solving}
Yang Song, Liyue Shen, Lei Xing, and Stefano Ermon.
\newblock Solving inverse problems in medical imaging with score-based
  generative models.
\newblock \emph{arXiv preprint arXiv:2111.08005}, 2021.

\bibitem[Sun et~al.(2020)Sun, Wang, Liu, Siegel, and Sarma]{sun2020pointgrow}
Yongbin Sun, Yue Wang, Ziwei Liu, Joshua Siegel, and Sanjay Sarma.
\newblock Pointgrow: Autoregressively learned point cloud generation with
  self-attention.
\newblock In \emph{Proceedings of the IEEE/CVF Winter Conference on
  Applications of Computer Vision}, pages 61--70, 2020.

\bibitem[Sutton et~al.(1999)Sutton, McAllester, Singh, and
  Mansour]{sutton1999policy}
Richard~S Sutton, David McAllester, Satinder Singh, and Yishay Mansour.
\newblock Policy gradient methods for reinforcement learning with function
  approximation.
\newblock \emph{Advances in neural information processing systems}, 12, 1999.

\bibitem[Tang et~al.(2022)Tang, Qian, Zhang, Zeng, Hou, and
  Zhe]{tang2022warpinggan}
Yingzhi Tang, Yue Qian, Qijian Zhang, Yiming Zeng, Junhui Hou, and Xuefei Zhe.
\newblock Warpinggan: Warping multiple uniform priors for adversarial 3d point
  cloud generation.
\newblock In \emph{Proceedings of the IEEE/CVF Conference on Computer Vision
  and Pattern Recognition}, pages 6397--6405, 2022.

\bibitem[Yang et~al.(2019)Yang, Huang, Hao, Liu, Belongie, and
  Hariharan]{yang2019pointflow}
Guandao Yang, Xun Huang, Zekun Hao, Ming-Yu Liu, Serge Belongie, and Bharath
  Hariharan.
\newblock Pointflow: 3d point cloud generation with continuous normalizing
  flows.
\newblock In \emph{Proceedings of the IEEE/CVF International Conference on
  Computer Vision}, pages 4541--4550, 2019.

\bibitem[Zhou et~al.(2021)Zhou, Du, and Wu]{zhou20213d}
Linqi Zhou, Yilun Du, and Jiajun Wu.
\newblock 3d shape generation and completion through point-voxel diffusion.
\newblock In \emph{Proceedings of the IEEE/CVF International Conference on
  Computer Vision}, pages 5826--5835, 2021.

\end{thebibliography}

%%%%%%%%%%%%%%%%%%%%%%%%%%%%%%%%%%%%%%%%%%%%%%%%%%%%%%%%%%%%

\appendix
\clearpage
\appendix

\section{Appendix I - Detailed Derivations}

\subsection{Derivation from Eq.~\eqref{eq:lossrl} to \eqref{eq:RL-kl}}\label{sec:A_RAML}\label{sec:eq7to11}
Below is the detailed derivation from Eq.~\eqref{eq:lossrl} to \eqref{eq:RL-kl}:
\begin{equation}\scriptsize
\begin{split}
& \loss_{RL}\\
= & - \E_{ \data} 
\bigg\{ 
\sum_{t=1}^T  \big[  \ent( p_\theta( \bX_{t-1} \mid \bX_{t} ) )  + \E_{ p_\theta(\bX_{t-1} \mid \bX_{t} ) } \reward( \bX_{t-1} )  \big] \bigg\} \\
= & - \E_{ \data} 
\bigg\{ 
\sum_{t=1}^T  \big[  \int p_\theta( \bX_{t-1} \mid \bX_{t} )  \log p_\theta( \bX_{t-1} \mid \bX_{t} )  \text{d} \bX_{t-1}  + \int p_\theta(\bX_{t-1} \mid \bX_{t} ) \reward( \bX_{t-1} ) \text{d} \bX_{t-1} \big] \bigg\} \\
\textcolor{black}{\bm{\propto}} & - \E_{ \data} 
\bigg\{ 
\sum_{t=1}^T  \big[  \int p_\theta( \bX_{t-1} \mid \bX_{t} ) \log p_\theta( \bX_{t-1} \mid \bX_{t} ) \text{d} \bX_{t-1}
-  p_\theta( \bX_{t-1} \mid \bX_{t} ) \big(  \reward( \bX_{t-1} ) - \underbrace{\log Z + \int q(\tilde{\bX}_{t-1} \mid \bX_t, \bX_0) \text{d} \tilde{\bX}_{t-1}  }_{Constants} \big) \text{d} \bX_{t-1} \\
\textcolor{black}{\bm{\propto}}  & - \E_{ \data} 
\bigg\{ 
\sum_{t=1}^T  \big[  \int p_\theta( \bX_{t-1} \mid \bX_{t} )  \log p_\theta( \bX_{t-1} \mid \bX_{t} )  \text{d} \bX_{t-1}  + \int p_\theta(\bX_{t-1} \mid \bX_{t} )   \int  p( \bX_{t-1} \mid \tilde{\bX}_{t-1} ) q(\tilde{\bX}_{t-1} \mid \bX_{t}, \bX_{0}) \text{d} \tilde{\bX}_{t-1}  \text{d} \bX_{t-1} \big] \bigg\} \\
= & \E_{ \data} 
\bigg\{ 
\sum_{t=1}^T \big[ \int p_\theta( \bX_{t-1} \mid \bX_t ) \bigg(  \log p_\theta( \bX_{t-1} \mid \bX_{t} ) - \log q_{pd}( \bX_{t-1} \mid \bX_t, \bX_0 ) \bigg ) \text{d} \bX_{t-1}  \big] \bigg\} \\
= & \E_{ \data } 
\sum_{t=1}^{T} \bigg\{ \big[ \kl( p_\theta( \bX_{t-1} \mid \bX_{t} ) \mid\mid  q_{pd}( \bX_{t-1} \mid \bX_t, \bX_0 ) )  \big] \bigg\},
\end{split}
\end{equation}
where $\int q(\tilde{\bX}_{t-1} \mid \bX_t, \bX_0) \text{d} \tilde{\bX}_{t-1}$ and $ \int p_\theta(\bX_{t-1} \mid \bX_{t} ) \text{d} \bX$ are constants.

\subsection{Derivation of Eq.~\eqref{eq:RMLE-kl-obj}}

Below is the detailed derivation of Eq.~\eqref{eq:RMLE-kl-obj}. Here, we introduce $\mathcal{F} (\bX_{t-1})$, which denotes the feasible set of $\tilde{\bX}_{t-1}$ given $\bX_{t-1}$. In this paper, $\tilde{\bX}_{t-1}$ lies within an $\ell_2$ ball of radius $d$ (as defined in Eq.~\eqref{eq:qpd}).
\begin{equation}\scriptsize
\begin{split}
& \loss_{RMLE} \\
= & \E
\bigg \{  \sum_{t=1}^{T} \int q_{pd}( \bX_{t-1} \mid \bX_t, \bX_0 ) \big(\log q_{pd}( \bX_{t-1} \mid \bX_t, \bX_0 ) - \log p_\theta( \bX_{t-1} \mid \bX_{t} ) \big) \text{d} \bX_{t-1} \bigg \} \\
= & \E
\bigg \{\sum_{t=1}^{T} \int \left[ 
\int_{\mathcal{F}(\bX_{t-1})} p(\bX_{t-1} \mid \tilde{\bX}_{t-1}) q ( \tilde{\bX}_{t-1} \mid \bX_t, \bX_0 ) \text{d} \tilde{\bX}_{t-1}
\bigg( \log \int_{\mathcal{F}(\bX_{t-1})} p(\bX_{t-1} \mid \tilde{\bX}_{t-1}) q ( \tilde{\bX}_{t-1} \mid \bX_t, \bX_0 ) \text{d}\tilde{\bX}_{t-1} \right.\\
& \left.\quad - ( \frac{1}{2}\log \beta_t - \frac{1}{2}\log \pi - \frac{1}{2}\log 2  - \frac{ \| \bX_{t-1} - \bmu_\theta(\bX_{t}, t) \|^2 }{2\beta_t} ) \bigg) \right]\text{d} \bX_{t-1} \bigg \} \qquad \text{  \textcolor{blue}{$ \Longleftarrow \frac{1}{Z} \exp( r (\bX_{t-1}) ) $ does not include $\tilde{\bX}_{t-1}$.} }  \\
= & \E
\bigg \{\sum_{t=1}^{T} \int \frac{1}{Z} \exp( r (\bX_{t-1}) ) \left[
\int_{\mathcal{F}(\bX_{t-1})} q ( \tilde{\bX}_{t-1} \mid \bX_t, \bX_0 ) \text{d} \tilde{\bX}_{t-1}
\bigg( \log \big( \frac{1}{Z} \exp( r (\bX_{t-1}) ) \int_{\mathcal{F}(\bX_{t-1})} q ( \tilde{\bX}_{t-1} \mid \bX_t, \bX_0 ) \text{d} \tilde{\bX}_{t-1} \big) \right. \\
& \left.\quad - ( \frac{1}{2}\log \beta_t - \frac{1}{2}\log \pi - \frac{1}{2}\log 2  - \frac{ \| \bX_{t-1} - \bmu_\theta(\bX_{t}, t) \|^2 }{2\beta_t} ) \bigg) \right]\text{d} \bX_{t-1} \bigg \} \qquad \text{ \textcolor{blue}{$\Longleftarrow$ finite integral of Gaussian family is a constant.} } \\
= & \E
\bigg \{\sum_{t=1}^{T} \int \frac{1}{Z} \exp( r (\bX_{t-1}) )
\bigg( \reward( \bX_{t-1} ) - \log Z  + \log \int_{\mathcal{F}(\bX_{t-1})} q ( \tilde{\bX}_{t-1} \mid \bX_t, \bX_0 ) \text{d} \tilde{\bX}_{t-1} \\
& \quad - ( \frac{1}{2}\log \beta_t - \frac{1}{2}\log \pi - \frac{1}{2}\log 2  - \frac{ \| \bX_{t-1} - \bmu_\theta(\bX_{t}, t) \|^2 }{2\beta_t} ) \bigg) \text{d} \bX_{t-1} \bigg \} \\
= & \E
\bigg \{\sum_{t=1}^{T}\int \frac{1}{Z} \exp( r (\bX_{t-1}) ) 
\bigg( constant + \frac{ \| \bX_{t-1} - \bmu_\theta(\bX_{t}, t) \|^2 }{2\beta_t} \bigg) \text{d} \bX_{t-1} \bigg \} \qquad \text{ \textcolor{blue}{ $\Longleftarrow$ the formulation of $\reward( \bX_{t-1} )$ is unknown. } } \\
\propto \; & \E
\bigg \{ 
\sum_{t=1}^{T} 
 \E_{\bX_t} \exp(  r (\bX_{t-1}) ) \frac{ \| \bX_{t-1} - \bmu_\theta(\bX_{t}, t) \|^2 }{2\beta_t} \bigg \}.
\end{split}
\end{equation}

\end{document}